\documentclass[conference,10pt]{IEEEtran} 
\IEEEoverridecommandlockouts
\usepackage{spconf,amsmath,graphicx}
\usepackage{endnotes}
\usepackage{epsfig,psfrag}
\usepackage{pst-all}
\usepackage{amsmath,amsthm,amssymb,amsfonts,upref,cite,epsf,color,bm}
\usepackage{graphicx}
\usepackage{color}
\usepackage{amsmath}
\usepackage{graphicx}
\usepackage{calc}
\usepackage{booktabs}
\usepackage{tikz}
\usepackage{pgfplots}
\newcommand\defeq{:=}

\usepackage{algorithm}
\usepackage{algpseudocode}
\floatname{algorithm}{Algorithm}
\algnewcommand\algorithmicinput{\textbf{Input:}}
\algnewcommand\INPUT{\item[\algorithmicinput]}
\algnewcommand\algorithmicoutput{\textbf{Output:}}
\algnewcommand\OUTPUT{\item[\algorithmicoutput]}
\algrenewcommand\algorithmicrequire{\textbf{Initialize:}}
\algnewcommand\INIT{\item[\algorithmicrequire]}

\DeclareMathOperator*{\argmin}{arg\;min}

\DeclareMathOperator*{\expect}{E}

\newcommand\vect[1]{\mathbf #1}

\newcommand{\vr}{\vect{r}}  
  
\newcommand{\vt}{\vect{t}}

\newcommand{\vx}{\vect{x}}  
\newcommand{\vy}{\vect{y}}  
\newcommand{\vz}{\vect{z}}

\newcommand{\measlen}{M}
\newcommand{\clusteridx}{r}
\newcommand{\numsteps}{L}
\newcommand{\nmse}{\hat{\varepsilon}}
\newcommand{\empnmse}{\bar{\varepsilon}}
\newcommand{\signalsize}{N}

\newcommand{\edges}{\mathcal{E}}
\newcommand{\cluster}{\mathcal{C}}
\newcommand{\nodes}{\mathcal{V}}
\newcommand{\graph}{\mathcal{G}}
\newcommand{\samplingset}{\mathcal{M}}

\DeclareMathOperator*{\prob}{P}
\newtheorem{theorem}{Theorem}
\newtheorem{lemma}[theorem]{Lemma}

\newcommand{\equref}[1]{(\ref{#1})}
\newcommand{\head}[1]{\textbf{#1}}
\usepackage{setspace}

\title{Random Walk Sampling For Big Data Over Networks}
\name{Saeed Basirian, Alexander Jung}
\address{\normalsize Department of Computer Science, Aalto University, Finland; firstname.lastname(at)aalto.fi\\[-0.5mm]
}

\begin{document}
	\maketitle
\begin{abstract}
It has been shown recently that graph signals with small total 
variation can be accurately recovered from only few samples if the 
sampling set satisfies a certain condition, referred to as the network 
nullspace property. Based on this recovery condition, we propose a 
sampling strategy for smooth graph signals based on random walks. 
Numerical experiments demonstrate the effectiveness of this approach 
for graph signals obtained from a synthetic random graph model 
as well as a real-world dataset. 
\end{abstract}

\begin{keywords} compressed sensing, big data, graph signal processing, total variation, complex networks
\end{keywords} 

\section{Introduction}
 \label{sec_intro}
 
Modern information processing systems are generating massive datasets which are partially 
labeled mixtures of different media (audio, video, text). Many successful approaches to such 
datasets are based on representing the data as networks or graphs. In particular, 
within (semi-)supervised machine learning, we represent the datasets by graph signals
defined over an underlying graph, which reflects the similarity relations between individual data points.
These graph signals often conform to a smoothness hypothesis, 
i.e., the signal values of close-by nodes are similar. 

Two key problems related to processing these datasets are (i) how to sample them, i.e., 
which nodes provide the most information about the entire dataset, and (ii) how to 
recover the entire graph signal representation of the dataset from these samples. 
These problems have been studied in \cite{Jung2016} which proposed a convex 
optimization method for recovering a graph signal from a small number of samples. 
Moreover, a sufficient condition for this recovery method to be accurate has been 
presented. This condition is a reformulation of the stable nullspace property of 
compressed sensing to the graph signal setting. 
 
{\bf Contribution.} Based on the intuition provided by the recently derived network 
nullspace property, we propose a sampling strategy based on random walks. 
The effectiveness of this approach is confirmed via numerical experiments based on 
synthetic graph signals obtained from a particular random graph model, i.e., 
the assortative planted partition model, and graph signals induced by a real-world 
dataset containing product rating information of an online retail shop.

{\bf Notation.} 
Vectors and matrices are denoted by boldface lower-case and upper-case letters, respectively. 
The vector with all entries equal to one (zero) is denoted $\mathbf{1}$ ($\mathbf{0}$). 
The $\ell_{1}$ and $\ell_{2}$ norm of a vector $\vx = (x_{1}, \ldots , x_{N})^{T}$ 
are denoted by $\|\vx\|_{1}$ and $\|\vx\|_{2}$ respectively.

{\bf Outline.} The problem setup is discussed in \ref{sec_setup}, were we formulate 
the problem of recovering a smooth graph signal as a convex optimization problem. 
Our main contribution is contained in Section \ref{sec_rws} where 
we present the random walk sampling method and discuss its properties in the context 
of the assortative planted partition model. The results of illustrative numerical experiments 
are presented in Section \ref{sec4_numerical}. We finally conclude in Section \ref{sec5_conclusion}. 

\section{Problem Formulation}
\label{sec_setup}

We consider massive heterogeneous datasets with intrinsic network structure 
represented by a  graph $\mathcal{G}=(\mathcal{V},\mathcal{E})$. 
The graph $\mathcal{G}$ consists of the nodes $\mathcal{V}=\{1,\ldots,\signalsize\}$, 
which are connected by undirected edges $\{i,j\} \in \mathcal{E}$. 
Each node $i\in \nodes$ represents an individual data point and an edge $\{i,j\} \in \edges$ 
connects nodes representing similar data points.  
For a given node $i \in \nodes$, we define its neighbourhood as 
\begin{equation}
\mathcal{N}(i) \defeq \{ j \in \mathcal{V}: \{i,j\} \in \mathcal{E} \}. 
\end{equation} 
The degree $d_{i} \defeq |\mathcal{N}(i)|$ of node $i\in \mathcal{V}$ counts the 
number of its neighbours. 

Within (semi-)supervised learning, we associate each data point $i\in \nodes$ with a label $x[i] \in \mathbb{R}$. 
These labels induce a graph signal $x[\cdot]: \nodes \rightarrow \mathbb{R}$ defined over the graph 
$\mathcal{G}$ underlying the dataset. 

We aim at recovering a smooth graph signal $\vx$ based on observing its values $x[i]$ 
for all nodes $i \in \mathcal{V}$ which belong to the sampling set 
\begin{equation} 
\mathcal{M}\defeq\{i_{1},\ldots,i_{\measlen}\} \subseteq \mathcal{V}.
\end{equation}
The size $\measlen \defeq |\mathcal{M}|$ of the sampling set is typically 
much smaller than the overall dataset, i.e., $\measlen \ll \signalsize$. For 
a fixed sampling budget $\measlen$ it is important to choose the sampling 
set such that the information obtained is sufficient to recover the overall 
graph signal. By considering a particular recovery method, 
called sparse label propagation (SLP), \cite{Jung2016} presents the network 
nullspace property as a sufficient condition on the sampling set 
such that SLP recovers the overall graph signal from the samples. 

The SLP recovery method is based on a smoothness hypothesis, which requires 
signal values of nodes belonging to the same cluster to be similar. 
This smoothness hypothesis then suggests to search for the particular graph signal 
which is consistent with the observed signal samples, and moreover has 
minimum total variation (TV) 
\begin{equation} 
\label{equ_def_TV}
\| \vx \|_{\rm TV} \defeq \sum_{\{i,j\} \in \mathcal{E}} |x[j]\!-\!x[i]|, 
\end{equation} 
which quantifies signal smoothness. 
Thus the recovery problem amounts to the convex optimization problem 
\begin{align}
\label{equ_optim_prob}
\hat{\vx} &\in \argmin \| \tilde{\vx} \|_{\rm TV} \quad {\rm s.t.} \quad \tilde{\vx}_{\mathcal{M}} = \vx_{\mathcal{M}}. 
\end{align}
The SLP algorithm is nothing but the the primal-dual optimization 
method of Pock and Chambolle \cite{pock_chambolle} applied to the problem \eqref{equ_optim_prob}. 

Let us from now on assume that the true underlying graph signal $\vx$ is clustered, i.e., 
\begin{equation}
\label{equ_clust_gsig}
\vx = \sum_{\mathcal{C} \in \mathcal{F}} a_{\mathcal{C}} \vt_{\mathcal{C}}, 
\end{equation}
with the cluster indicator signals
\begin{equation} 
\label{equ_def_indicator_signal}
t_{\mathcal{C}}[i] = \begin{cases} 1 \mbox{, if }  i \in \mathcal{C} \\ 0 \mbox{ else.}  \end{cases}
\end{equation}
For a partition $\mathcal{F}=\{\mathcal{C}_{1},\ldots,\mathcal{C}_{|\mathcal{F}|}\}$ consisting 
of disjoint clusters $\mathcal{C}_{l}$ with small cut-sizes, we have that the TV $\| \vx \|_{\rm TV}$ 
is relatively small. Thus, we expect recovery based on TV minimization \eqref{equ_optim_prob} 
to be accurate for signals of the type \eqref{equ_clust_gsig}. Indeed, a sufficient condition 
for the solution $\hat{\vx}$ of \eqref{equ_optim_prob} to coincide with 
$\vx= \sum_{\mathcal{C} \in \mathcal{F}} a_{\mathcal{C}} \vt_{\mathcal{C}}$ can be 
formulated as  
\begin{lemma}
\label{lem_suff_cond_NNSP}
We observe a clustered signal $\vx$ of the form \eqref{equ_clust_gsig} 
on the sampling set $\samplingset \!\subseteq\! \nodes$. If each boundary edge 
$\{i,j\}$ with  $i\!\in\!\mathcal{C}_{a}$, $j \!\in\! \mathcal{C}_{b}$ is 
connected to two sampled nodes in each cluster, i.e., 
\begin{equation} 
|\samplingset \cap \mathcal{C}_{a} \cap \mathcal{N}(i)|\geq2 \mbox{, and } |\samplingset \cap \mathcal{C}_{b} \cap \mathcal{N}(j)|\geq2,
\end{equation}
then \eqref{equ_optim_prob} has a unique solution which moreover coincides with the true 
graph signal $\vx$. 
\end{lemma} 

\section{Random Walk Sampling} 
\label{sec_rws}
We now present a particular strategy (summarized in Algorithm \ref{alg_RWS} below) 
for choosing the sampling set $\samplingset$ of nodes at which the graph signal should 
be sampled to obtain the observations $\{x[i]\}_{i \in \samplingset}$.  
Our strategy is based on parallel random walks which are started at randomly 
selected seed nodes. The endpoints of these random walks, which are run for 
a fixed number $L$ of steps, constitute the sampling set $\samplingset$.

\begin{algorithm}
	\caption{Random Walk Sampling}
	\label{alg_RWS}
	\begin{algorithmic}[1]
		\INPUT{random walk length $L$, sample budget $M$}
		\INIT{Sampling set $\mathcal{M} = \emptyset$}
		\For{$j = 1:M$}  
		\State randomly select a seed (start) node $i_{1}$
		\State perform a length-$L$ random walk $\mathcal{P}_{j} \leftarrow (i_{1}, \ldots , i_{L})$, 
		\State $\mathcal{M} \leftarrow \mathcal{M} \bigcup \{i_{j}\}$ 
		\EndFor
		\OUTPUT{$\mathcal{M}$}
	\end{algorithmic}
\end{algorithm} 
\begin{figure}
\hspace*{0.5em}\includegraphics[width=0.9\columnwidth]{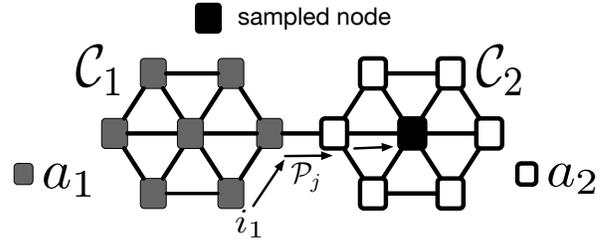}
\caption{\label{fig_paths}Clustered graph signal \eqref{equ_clust_gsig} defined over a graph composed of two clusters $\mathcal{C}_{1}$ and $\mathcal{C}_{2}$.}
\end{figure}
In Figure \ref{fig_paths}. we illustrate the construction of the sampling set via 
the random walks $\mathcal{P}_{j}$. Each random walk $\mathcal{P}_{j}$ 
forms a finite sequence $\{v_{1}=r_{j},\ldots,v_{\numsteps}=i_{j}\}$ 
of nodes that are visited in successive steps of the walk.

The sampling strategy of Algorithm \ref{alg_RWS} is appealing since it allows for efficient implementation 
as the random walks can be follows in parallel. Moreover, for a particular 
random graph model, the sampling set $\samplingset$ delivered by Algorithm \ref{alg_RWS} conforms 
with Lemma \ref{lem_suff_cond_NNSP}. According to Lemma \ref{lem_suff_cond_NNSP}, we have to 
select from each cluster $\mathcal{C}_{l}$ a number sampled nodes which is 
proportional to its cut-size $|\partial \mathcal{C}_{l}|$. Thus, 
we have to sample more densely in those clusters which have large cut-size. We now 
show that the sampling set $\samplingset$ obtained by Algorithm \ref{alg_RWS} follows this rationale for 
graph signals obtained from the stochastic block model (SBM) \cite{Mossel2012}.

For a given partition $\mathcal{F}=\{\mathcal{C}_{1},\ldots,\mathcal{C}_{|\mathcal{F}|}\}$ of the graph $\graph$ in clusters $\mathcal{C}_{l}$ 
of size $\signalsize_{l} \defeq |\mathcal{C}_{l}|$, the SBM 
is a generative stochastic model for the edge set $\edges$ of the graph $\graph$. In its simplest 
form, which is called the assortative planted partition model (APPM)\cite{Mossel2012}, the SBM 
is defined by two parameters $p$ and $q$ which specify the probability that two particular 
nodes $i,j$ of the graph are connected by an edge $\{i,j\}$. In particular, two nodes $i,j \in \mathcal{C}_{i}$ 
out of the same cluster are connected by an edge with probability $p$, i.e., $\prob\{ \{i,j\} \in \edges \} = p$ for $i,j \in \mathcal{C}_{a}$. 
Two nodes $i \in \mathcal{C}_{a}$, $j \in \mathcal{C}_{b}$ from different clusters $\mathcal{C}_{a}$ and $\mathcal{C}_{b}$ 
are connected by an edge with probability $q$, i.e., $\prob\{ \{i,j\} \in \edges \} = q$ for $i \in \mathcal{C}_{a}$ and $j \in \mathcal{C}_{b}$. 

Elementary derivations yield the expected degree $\bar{d}_{\clusteridx}$ of any 
node $i\in \mathcal{C}_{\clusteridx}$ belonging to cluster $\mathcal{C}_{\clusteridx}$ as 
\begin{equation}
\label{equ_mean_degree}
\bar{d}_{\clusteridx} = \expect \{ d_{i} \} = p(\signalsize_{\clusteridx}-1) + q (\signalsize-\signalsize_{\clusteridx}). 
\end{equation}
On the other hand, by similarly elementary calculations, the 
expected cut-size $\overline{C}_{\clusteridx} \defeq |\partial \mathcal{C}_{\clusteridx}|$ satisfies 
\begin{equation}
\label{equ_expected_cluster_size}
\overline{C}_{\clusteridx} = q \signalsize_{\clusteridx} (\signalsize - \signalsize_{\clusteridx}). 
\end{equation}

Now consider a particular random walk $\mathcal{P}_{j}$ which is run in Algorithm \ref{alg_RWS}. 
For a fixed node $i \in \nodes$, let $p_{l}(i)$ denote the probability that the random walk 
visits node $i$ in the $l$th step. 
A fundamental result in the theory of random walks over graphs 
states \cite[page 159]{Newman2010}
\begin{equation}
\label{equ_rw_stat_dist}
\lim_{l \rightarrow \infty} p_{l}(i) = \frac{d_{i}}{2|\edges|}
\end{equation}
Thus, by running the random walks in Algorithm \ref{alg_RWS} sufficiently long (choosing $L$ sufficiently large), 
the probability that the delivered sampling set $\samplingset$ contains a node $i \in \cluster_{\clusteridx}$ from 
cluster $\mathcal{C}_{\clusteridx}$ statisfies 
\begin{equation}
\label{equ_includes_samplingset}
\prob \{ i \in \samplingset \} \approx  \frac{p(\signalsize_{\clusteridx}-1) + q (\signalsize-\signalsize_{\clusteridx})}{2|\edges|}.
\end{equation} 
Contrasting \eqref{equ_includes_samplingset} with \eqref{equ_expected_cluster_size} reveals that 
the sampling set delivered by Algorithm \ref{alg_RWS} indeed conforms with Lemma \ref{lem_suff_cond_NNSP}, which requires 
clusters with larger cut-size to be sampled more densely.

\section{Numerical Results} 
\label{sec4_numerical}
We tested the effectiveness of the sampling method given by Algorithm \ref{alg_RWS} was 
verified by applying it to different graph signals and using sparse label propagation (SLP) as the 
recovery method for obtaining the original graph signal from the samples. 
The SLP algorithm, derived in \cite{Jung2016}, is restated as Algorithm \ref{alg_SLP} 
for convenience. In Algorithm \ref{alg_SLP}, we make use of the clipping operator 
$\mathcal{T} : \mathbb{R}^{|\mathcal{E}|} \rightarrow \mathbb{R}^{|\mathcal{E}|}$ 
for edge signals defined element-wise as $(\mathcal{T}(\tilde{\vx}))[e] = (1/\text{max}\{|\tilde{\vx}[e]|, 1\})\tilde{\vx}[e]$. 

\begin{algorithm}
	\caption{Sparse Label Propagation \cite{Jung2016}}
	\label{alg_SLP}
	\begin{algorithmic}[1]
		\INPUT{data graph $\graph$, sampling set $\samplingset$, signal samples $\{x[i]\}_{i \in \mathcal{M}}$}.
		\INIT{$k\!\defeq\!0$, $\mathbf{D} \defeq$ incidence matrix of $\graph$ for some arbitrary orientation, $\vz^{(0)}\!\defeq\!\mathbf{0}, \vx^{(0)}\!\defeq\!\mathbf{0}, \hat{\vx}^{(0)}\!\defeq\!\mathbf{0}, \vy^{(0)}\!\defeq\!\mathbf{0}$, maximum node degree $d_{\rm max}\defeq \max_{i \in \nodes} d_{i}$}
		\Repeat
		\State $\vy^{(k+1)} := \mathbf{\mathcal{T}}(\vy^{(k)}+(1/2\sqrt{d_{\rm max}})\mathbf{D}\vz^{(k)})$
		\State $\vr := \vx^{(k)} - (1/2\sqrt{d_{\rm max}})\mathbf{D}^{T}\vy^{(k+1)}$
		\State $\vx^{(k+1)} := \begin{cases} x[i] & \mbox{ for } i \in \mathcal{M} \\ 
		r[i] & \mbox{ else. } \end{cases}$
		\State $\vz^{(k+1)} := 2\vx^{(k+1)} - \vx^{(k)}$
		\State $\hat{\vx}^{(k+1)} := \hat{\vx}^{(k)} + \vx^{(k+1)}$
		\State $k := k + 1$
		\Until{stopping criterion is satisfied}
		\OUTPUT{$\hat{\vx}^{(k)} := (1/k)\hat{\vx}^{(k)}$}
	\end{algorithmic}
\end{algorithm}

\begin{figure}
	\centering
	\hspace*{0.5em}\includegraphics[width = \columnwidth, height= \columnwidth]{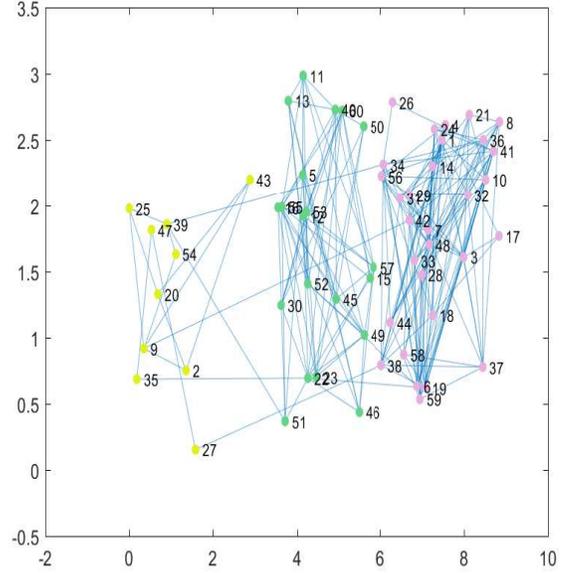}
	\caption{An APPM instance with 60 nodes and three clusters. Node colours represent the signal values.}\label{fig_SBM}
\end{figure}

Our numerical experiments involved $10^4$ independent simulation runs. 
Each simulation run is based on randomly generating an instance (see Figure \ref{fig_SBM}) 
of the APPM for fixed parameter values  $p=3/10$, $q=5/100$ and 
partition consisting of four clusters with sizes $|\mathcal{C}_{1}|=10, 
|\mathcal{C}_{2}|=20, |\mathcal{C}_{3}|=30, |\mathcal{C}_{4}|=40$ ((cf.\ Section \ref{sec_rws}). 
We then generated a clustered graph signal $\vx$ of the form \eqref{equ_clust_gsig} 
by choosing the cluster values $a_{\mathcal{C}}$ as independent 
random variables $a_{\mathcal{C}} \sim U(0, 1) \ (\text{cf.} \ \equref{equ_clust_gsig})$ .
  
For each realization of the APPM, we constructed a sampling set $\samplingset$ 
using Algorithm \ref{alg_RWS} which was then used to obtain the signal samples 
$\{x[i]\}_{i \in \samplingset}$ and subsequently recovering the entire graph signal 
$\vx$ via Algorithm \ref{alg_SLP}. 
We measured the recovery accuracy obtained by Algorithm \ref{alg_SLP} via 
the normalized empirical mean squared error (NMSE) of the signal estimate $\hat{\vx}$, i.e., 
\begin{equation} 
\nmse^{(l)} \defeq \frac{\| \hat{\vx}^{(l)} - \vx^{(l)} \|_{2}^{2}}{\| \vx^{(l)} \|_{2}^{2}}.
\end{equation} 
Here, $\nmse^{(l)}$, $\vx^{(l)}$ and $\hat{\vx}^{(l)}$ denote the NMSE, the original and the 
recovered graph signal, respectively, obtained in the $l$th simulation run. 
Note that $\nmse$ is random and often we are interested in its empirical mean 
\begin{equation}
\empnmse \defeq (1/10^4) \sum_{l=1}^{10^{4}} \nmse^{(l)}. 
\end{equation}

We evaluated the quality of the sampling set provided by Algorithm \ref{alg_RWS} for varying 
sampling budgets $\measlen$ and a fixed length $L=10$ of the random 
walks $\mathcal{P}_{j}$. In Table \ref{Table 1}, we report the mean and standard deviation 
of the NMSE of $\hat{\vx}$ for different sampling budgets $\measlen$. 
Besides the expected decrease in error by increasing the number of samples, 
it shows that sampling around half of graph 
nodes, we obtain $\nmse \approx 0.082$. 

We also investigated the effect of choosing a varying random walk length $L$ in Algorithm \ref{alg_RWS}, 
for a fixed sample budget $\measlen=10$. In Table \ref{Table 2}, we display the mean and 
standard deviation of the NMSE for different values of $L$. It shows that for these range of 
values, the length of the walks have a relatively insignificant effect on the outcome. 
This can be partially explained by the fact that the mixing time of random walks 
(i.e., the number of steps before they reach the stationary distribution) 
in some cases may be much less than the size of the graph $\signalsize$ \cite{Lovasz93}.

\begin{table}
	\centering
	\begin{tabular}{@{}l*5c@{}}
		\toprule[1.5pt]
		&\multicolumn{5}{c}{\head{Sampling Budget $\measlen$}}\\
		&\measlen=10 & 20 & 30 & 40 & 50\\
		\cmidrule(lr){2-6}
		$\empnmse$ & 0.285 & 0.232 & 0.188 & 0.132 & 0.082\\
		STD & 0.221 & 0.178 & 0.160 & 0.138 & 0.091\\
		\bottomrule[1.5pt] 
	\end{tabular}
	\caption{Average NMSE $\empnmse$ obtained for different sampling budgets $\measlen$. STD indicates the empirical 
	standard deviation of the NMSE $\nmse$.}\label{Table 1}
\end{table}

\begin{table}
	\centering
	\begin{tabular}{@{}l*5c@{}}
		\toprule[1.5pt]
		&\multicolumn{5}{c}{\head{Random Walk Length $L$}}\\
		&L=20 & 40 & 80 & 160 & 320\\
		\cmidrule(lr){2-6}
		$\empnmse$ & 0.312 & 0.314 & 0.285 & 0.277 & 0.304\\
		STD & 0.235 & 0.248 & 0.214 & 0.232 & 0.216\\
		\bottomrule[1.5pt] 
	\end{tabular}
	\caption{Average NMSE $\empnmse$ obtained for different lengths $L$ of the random walks. 
	STD indicates the empirical standard deviation of the NMSE $\nmse$.}\label{Table 2}
\end{table}

The fluctuation of the NMSE, as indicated by the values of the empirical standard deviation 
in Tables \ref{Table 1} and \ref{Table 2}, are on the order of the average NMSE. We expect 
the reason for this rather large amount of fluctuation to be a too small number of simulation runs. 
However, due to resource constraints we have not been able to increase the number of runs significantly. 

\begin{figure}
	\centering
	\hspace*{0.5em}\includegraphics[width = \columnwidth, height=8cm]{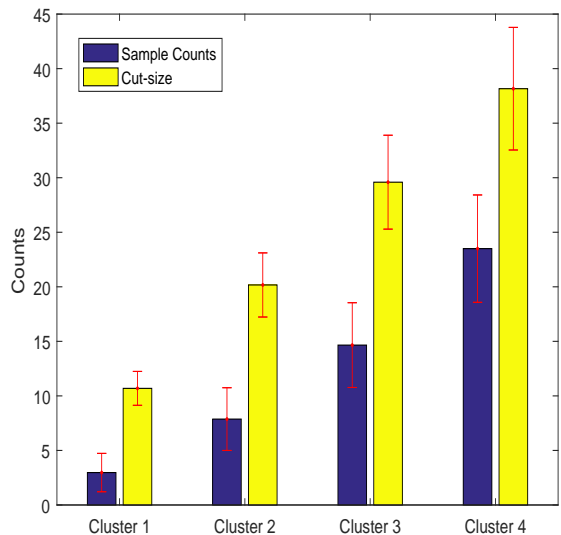}
	\caption{The mean number of samples and the mean cut-size of each cluster. $M = 50, L = 10$.}\label{fig_SampleCounts_L10}
\end{figure}

In the final experiment, we challenged the hypothesis that the sampling strategy 
conforms to the intuition, suggested by Lemma \ref{lem_suff_cond_NNSP}, of taking 
more samples in clusters with larger cut-size (cf.\ Section \ref{sec_rws}). 
For this purpose, the same procedure in the first two tests was repeated for $L=10$ 
and $\measlen=50$, and the number of samples in each cluster and its cut-size was 
recorded in each run. In Figure \ref{fig_SampleCounts_L10}, we report the obtained results, 
which indicates that the mean sample counts $|\samplingset \cap \cluster_{\clusteridx}|$ 
are approximately proportional to the cluster cut-sizes $| \partial \mathcal{C}_{\clusteridx}|$. 

\subsection{Real-World Data Set}
We also tested our approach on the Amazon co-purchase dataset from the 
Stanford Network Analysis Platform \cite{snapnets}. The dataset consists of a 
collection of products purchased on the Amazon website. For each product, 
it provides a list of other products that are frequently co-purchased with it, 
as well as an average user rating. We first extracted an undirected graph 
underlying the full dataset (excluding nodes with no co-purchase information), 
which includes an edge $\{i,j\}$ if product $j$ is co-purchased with product $i$ 
or vice versa. Subsequently, we selected a subgraph via a random walk and 
including all the nodes on the path and their neighbours, resulting in a graph 
with $N=5227$ nodes and $12758$ edges. The graph signal is the average 
user rating for the products.

The sampling set was extracted using the random walk method with the sampling 
ratio $M/N = 0.1$ and $L=20$. The SLP algorithm was then applied for recovering 
the graph signal. This resulted in a mean NMSE of $0.332 \pm 0.013$ over 10 runs. 
For comparison, we also tested three graph clustering algorithms (also referred to as 
community detection algorithms) for selecting the sampling set. This comprised of first 
finding the partitioning of the nodes using the clustering algorithms and then randomly 
sampling from each cluster, where the number of samples in clusters was uniformly 
distributed according to the cut-size. For finding the clusters, we used an algorithm by 
Blondel et. al. (also known as Louvain) \cite{blondel2008fast}, an algorithm by 
Newman \cite{newman2004fast}, and one by Ronhovde et. al. \cite{ronhovde2010local}. 
Choosing the sampling set via these methods and applying SLP for recovering the 
graph signal resulted in a NMSE of 0.369, 0.478, and 0.364 for the Louvain, Newman, 
and Ronhovde methods respectively (the value for the Ronhovde method is the 
average over 5 different clusterings corresponding to 5 values of its gamma 
parameter equally spaced between 0.1 and 0.5). We conclude that in this case our 
random walk method performs similarly to more computationally demanding 
clustering algorithms for sampling the graph signal.

\section{Conclusions}
\label{sec5_conclusion}
We proposed a novel random walk strategy for sampling graph signals representing 
massive datasets with intrinsic network structure. This strategy conforms with the rationale, 
which is supported by the recently derived network nullspace property, to sample more 
densely in clusters with large cut-size. The proposed sampling method has been tested on synthetic graph signals 
generated via an APPM. Our numerical experiments demonstrated that combining 
our sampling strategy with the SLP recovery algorithm, it is possible to recover graph signals 
with small error from only few samples. The effectiveness of our sampling strategy has been 
also verified numerically for graph signals obtained from a real-world dataset containing 
product rating information of an online retail shop. 

\vspace*{1mm}
\bibliographystyle{abbrv}
\bibliography{ListSAMPTA2017}
\end{document}